\documentclass[letterpaper, 10pt, conference]{ieeeconf} 
\usepackage[utf8]{inputenc}
\usepackage{amsmath}
\usepackage{amsfonts}
\usepackage{amssymb}
\usepackage{booktabs}
\usepackage{graphicx}
\usepackage{multirow}
\usepackage{subcaption}
\usepackage{url}
\usepackage{siunitx}
\usepackage{xcolor}
\usepackage{grffile}
\usepackage{tikz}

\newcommand{\nkeyframes}{n}
\newcommand{\minoverlapthreshold}{\tau}
\newcommand{\rangetolerance}{T_r}
\newcommand{\neighborhoodslicesize}{T_\alpha}
\newcommand{\invalidityratio}{r_\text{invalidity}}
\newcommand{\invalidityratiothreshold}{t_\text{unmerge}}

\IEEEoverridecommandlockouts
\title{\LARGE \bf Detecting Invalid Map Merges in Lifelong SLAM\vspace{-5mm}}
\author{Matthias Holoch\textsuperscript{\textsection}, Gerhard Kurz\textsuperscript{\textsection}, Peter Biber
	\thanks{The authors are with Robert Bosch GmbH, Corporate Research, Germany. E-mail: \{matthias.holoch, gerhard2.kurz, peter.biber\}@de.bosch.com%
	}
	\vspace{-2cm}
}

\newcommand\copyrighttext{%
	\footnotesize \textcopyright 2022 IEEE. Personal use of this material is permitted.
	Permission from IEEE must be obtained for all other uses, in any current or future
	media, including reprinting/republishing this material for advertising or promotional
	purposes, creating new collective works, for resale or redistribution to servers or
	lists, or reuse of any copyrighted component of this work in other works.
}
\newcommand\copyrightnotice{%
	\begin{tikzpicture}[remember picture,overlay]
	\node[anchor=south,yshift=10pt] at (current page.south) {\fbox{\parbox{\dimexpr\textwidth-\fboxsep-\fboxrule\relax}{\copyrighttext}}};
	\end{tikzpicture}%
}

\begin{document}
\maketitle
\copyrightnotice
\begingroup\renewcommand\thefootnote{\textsection}
\footnotetext{Equal contribution}
\endgroup

\begin{abstract}
  For Lifelong SLAM, one has to deal with temporary localization failures, e.g., induced by kidnapping.
  We achieve this by starting a new map and merging it with the previous map as soon as relocalization succeeds.
  Since relocalization methods are fallible, it can happen that such a merge is invalid, e.g., due to perceptual aliasing.
  To address this issue, we propose methods to detect and undo invalid merges.
  These methods compare incoming scans with scans that were previously merged into the current map and consider how well they agree with each other.
  Evaluation of our methods takes place using a dataset that consists of multiple flat and office environments, as well as the public MIT Stata Center dataset.
  We show that methods based on a change detection algorithm and on comparison of gridmaps perform well in both environments and can be run in real-time with a reasonable computational cost.
\end{abstract}

\section{Introduction}
Lifelong SLAM (see \cite[Sec.~III]{Cadena2016}) involves the lifelong operation of a robot that is moving through its environment while it is always mapping to extend and update the map.
Over the course of its lifetime, it will inevitably happen that localization is lost, e.g, when the robot is kidnapped (turned off, moved by a human, and turned on again), when sensors temporarily provide no or unusable data, or when the SLAM algorithm fails.
In such cases, the robot first starts building a new map and later performs relocalization (see \cite{Colosi2019,Bokovoy2020}) as soon as it can find its pose again with respect to the previously created map.
In this case, we merge the previous map with the current map to create a combined map that covers the entire area explored so far.
However, relocalization algorithms can fail under certain circumstances, e.g, due to perceptual aliasing when two different places in the map look alike.
In order to recover from this case, we propose an \emph{invalid merge detection} algorithm, which can later detect whether or not a merge was correct and, if the merge turns out to be incorrect, revert this merge.

\begin{figure}
	\centering
	\begin{subfigure}{\linewidth}
		\centering
		\fbox{
			\includegraphics[width=.43\linewidth]{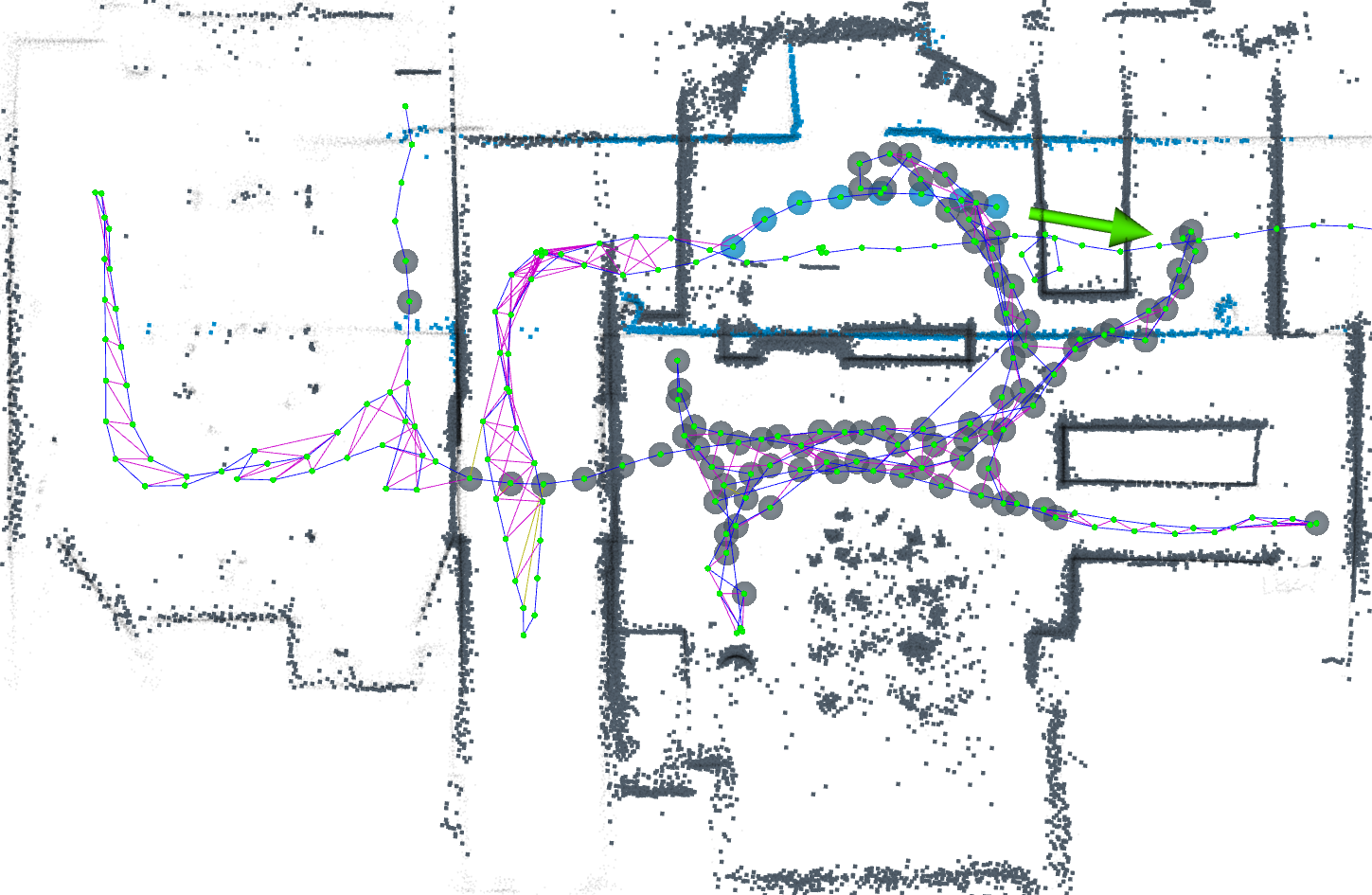}
		}
		\fbox{
			\includegraphics[width=.43\linewidth]{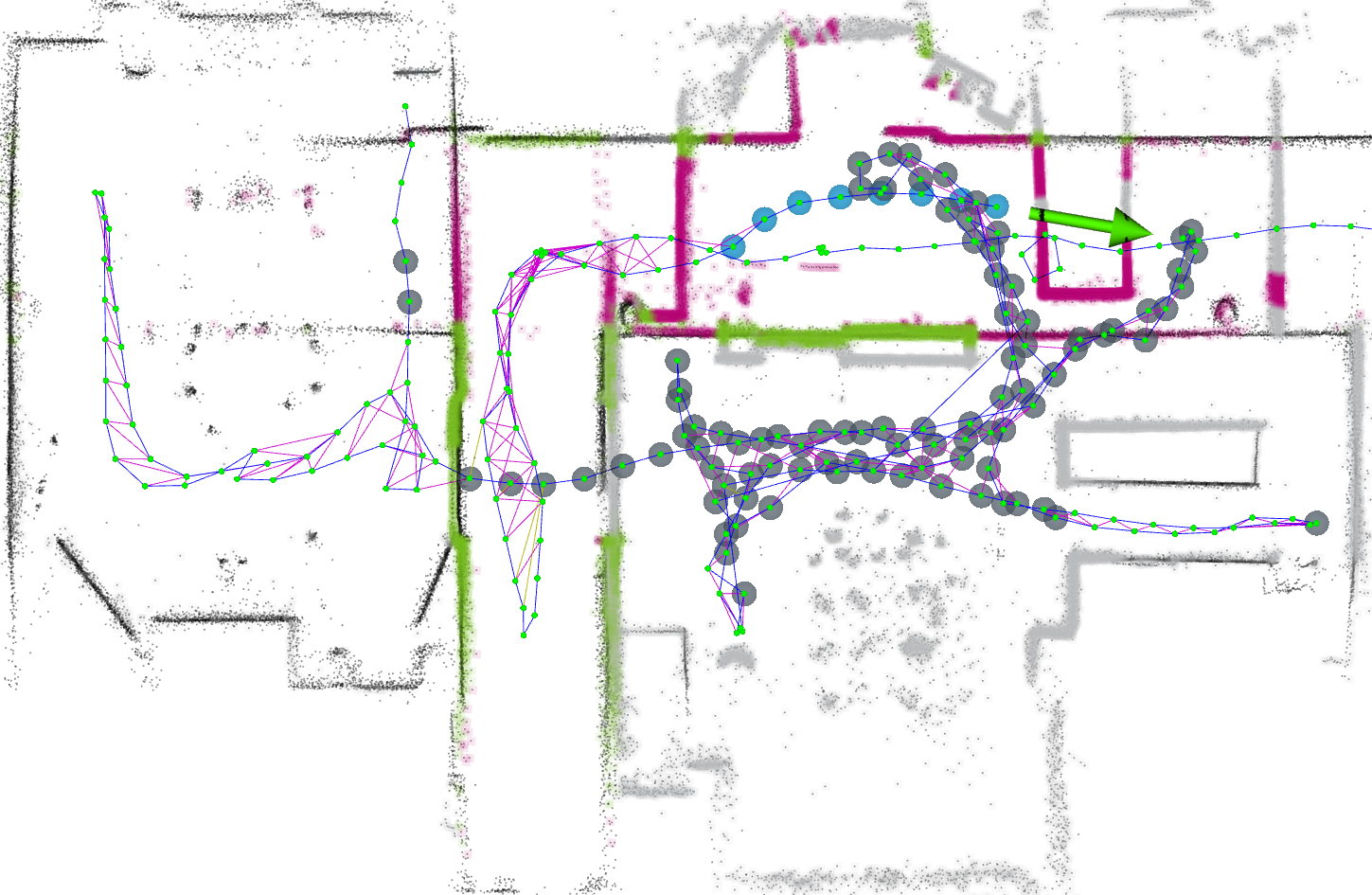}
		}
		\caption{Invalid merge.}
		\label{fig:into_example_invalid_merge}
	\end{subfigure}
	\begin{subfigure}{\linewidth}
		\centering
		\fbox{
			\includegraphics[width=.43\linewidth]{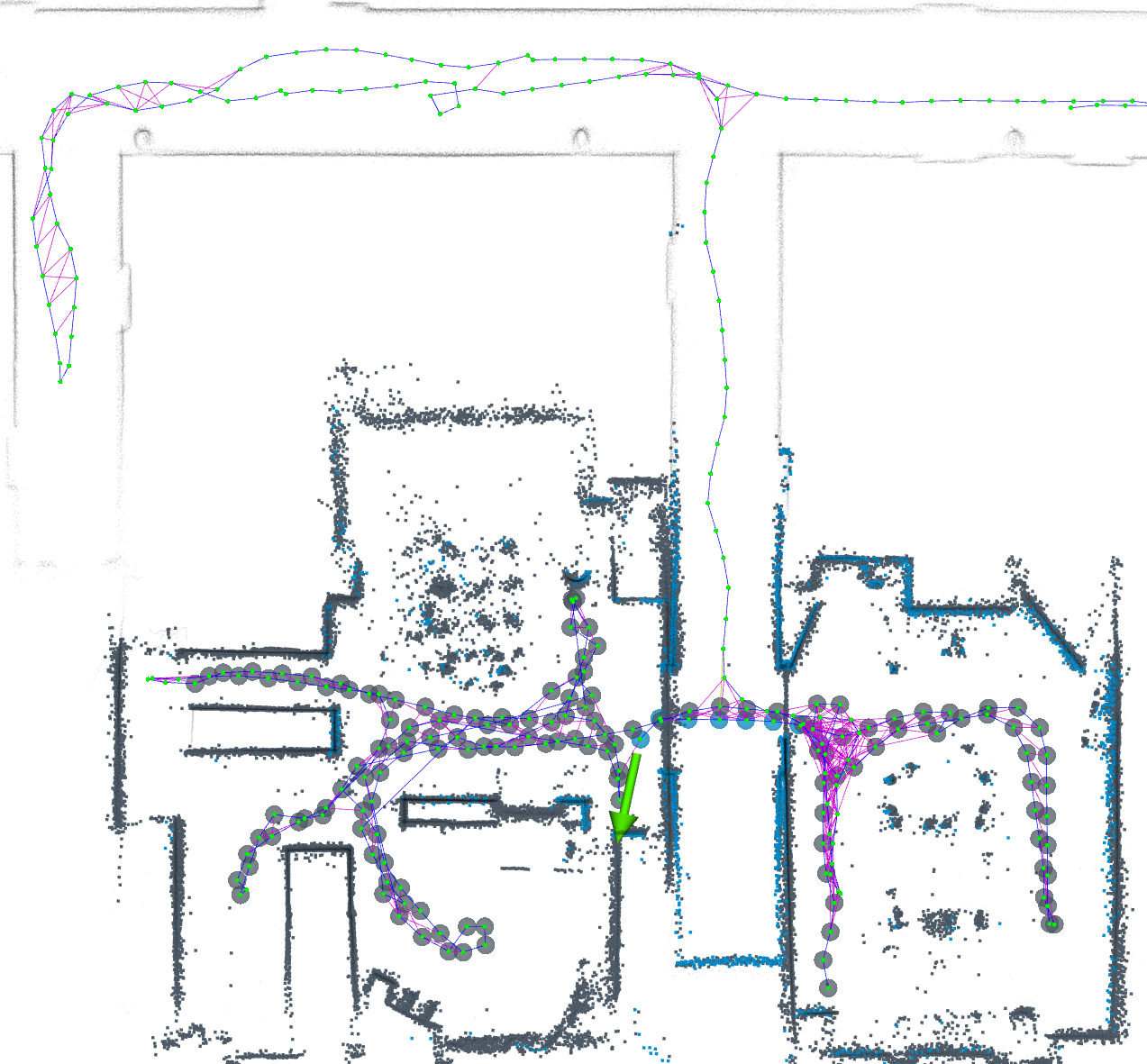}
		}
		\fbox{
			\includegraphics[width=.43\linewidth]{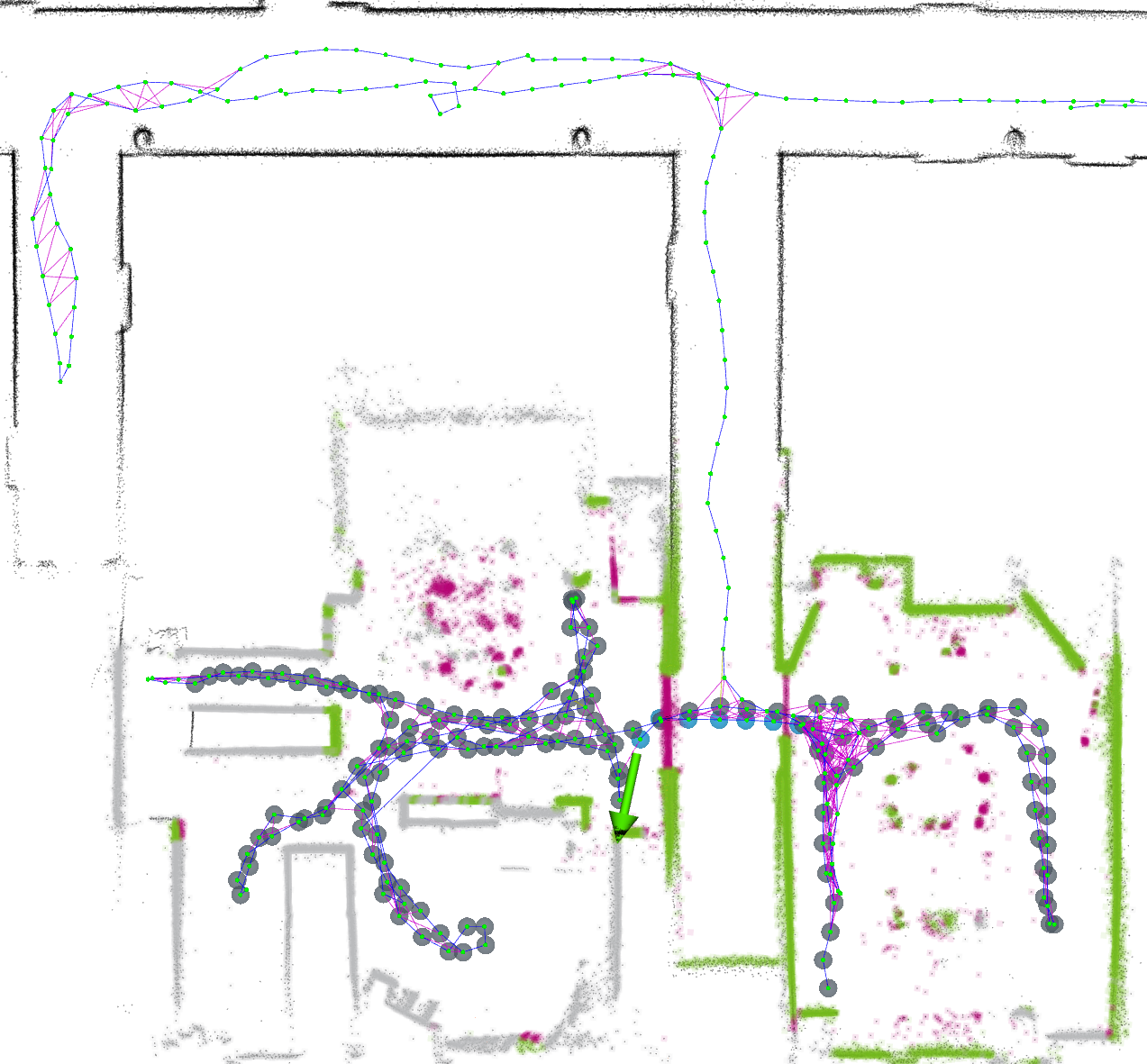}
		}
		\caption{Corrected merge.}
		\label{fig:into_example_corrected_merge}
	\end{subfigure}
	\caption{
		Example of an invalid map merge, due to an ambiguous corridor.
		Blue \textcolor[rgb]{0.00,0.56,0.81}{$\blacksquare$} circles are considered vertices from the active epoch, and the left images highlight the scan points that belong to them.
		Gray \textcolor[rgb]{0.32,0.37,0.41}{$\blacksquare$} corresponds to data from other epochs.
		The right images show the classification results from our change detection based method for detecting invalid merges.
		Red \textcolor[rgb]{0.698,0.145,0.298}{$\blacksquare$} corresponds to \emph{change}, green \textcolor[rgb]{0.596,0.686,0.239}{$\blacksquare$} to \emph{agree}, and gray \textcolor[rgb]{0.714,0.714,0.722}{$\blacksquare$} to \emph{no info}.
	}
	\label{fig:intro_example}
\end{figure}

Most approaches in literature that perform relocalization try to optimize their algorithms to achieve a high  relocalization accuracy.
However, it is usually not considered that even an algorithm with a high accuracy will eventually fail, if it runs for a long time and on a variety of datasets.
In particular, there are sometimes environments where two different places appear exactly the same to the robot and this ambiguity may not even be detectable by looking at the existing map because so far only one of these two places has been mapped.
In general, there is always a trade-off between the desire to relocalize quickly and the desire to relocalize reliably:
The longer one waits, the more information can be gathered, which lowers the chance of wrong relocalization.
However, early relocalization is desired because data from a previous map may be needed to allow the robot to continue its task (e.g. navigation to places in the other map).
Thus, we want to relocalize early, and be able to undo potential mistakes later.

When implementing an this idea, we are confronted with two main questions:
How to detect an invalid merge?
And how to undo that invalid merge?
In this paper, we provide answers to these questions with the following contributions:
\begin{itemize}
	\item Theoretical investigation into how invalid merges can be detected.
	\item Four concrete algorithms to efficiently detect invalid merges with high accuracy.
	\item A model for handling multiple, disconnected graphs in a graph-based SLAM system.
	\item Algorithm for merging graphs and undoing merges.
	\item Evaluation scheme based on a reference map that reduces manual effort to a minimum.
	\item Evaluation on two real-world, indoor datasets.
\end{itemize}

In the following, we assume a graph-based SLAM algorithm~\cite{Grisetti2010a} based on 2D lidar data.
Some of the methods proposed in this paper can also be generalized to the 3D case, however.

\section{Related Work}
Map merging has been studied mostly for multi-robot use cases where several robots map an environment together and try create a complete map by merging their individual maps~\cite{Yu2020}.
Due to the distributed nature of this application, the focus is often on the question of how the amount of data exchanged between the robots can be kept small~\cite{Schuster2018} and when it is the best time to merge~\cite{Dinnissen2012}.

However, some authors have also considered map merging for single-robot applications, where merging is used as a means to facilitate relocalization~\cite{Qin2018, Bokovoy2020}.
This is the use case of map merging on which we focus in this paper.

When it comes to map merging algorithms, different types of maps have been considered.
One common approach is to perform map merges on the level of a gridmap~\cite{Yu2020,Hernandez2020,Carpin2008,Liu2013}.
While merging gridmaps may be sufficient for some applications, this kind of merge is fairly limited because it cannot easily perform a non-rigid alignment where parts of the map are warped slightly.
In the case of graph-based SLAM approaches, the merged gridmap also has the disadvantage that it does not contain any graph structure anymore, so individual lidar scan poses cannot be optimized further when new data becomes available.
There is also some research on merging feature maps~\cite{Yu2020} and point clouds~\cite{Bokovoy2020}.
However, these strategies suffer from similar limitations.
Thus, we focus on merging maps at the level of the SLAM graph~\cite{Qin2018,Schuster2018} in this paper.

When it comes to detecting and undoing invalid map merges, we are not aware of any directly comparable research.
However, there have been some works on detecting wrong loop closures~\cite{Corso2013,Latif2013a,Lee2013}, which is a closely related problem, since an invalid map merge is fairly similar to a wrong global loop closure.
Furthermore there is some research on map quality measures~\cite{Almqvist2017,Droeschel2018,Filatov2017}, which could be used to detect invalid merges by computing a map quality measure and assuming that a invalid merge must have happened if it drops below a certain threshold.
We adapt the method in~\cite{Corso2013} (histogram) and~\cite{Droeschel2018} (entropy) to compare our methods against.

\section{Integration in SLAM Pipeline}
\label{sec:integration}
\newcommand{\epochid}{\iota}
\newcommand{\epochidset}{I}

In this section, we explain how to integrate map merging and unmerging into a graph-based SLAM pipeline.
First, we introduce the concept of an epoch: An epoch ends, whenever incremental localization methods get disrupted.
This is bound to happen, e.g., when the robot is turned off or detects that it was kidnapped.
As soon as normal operation resumes, a new epoch is started.
We assign each epoch a unique id $\epochid \in \mathbb{N}$ and we add it to each vertex $v$ that is submitted into a graph using data from that epoch.

Similar to~\cite{Colosi2019}, our architecture assumes that at any given point in time, there is a single active SLAM graph $G_{\epochidset^{\ast}_{n}}$ and a set of zero or more inactive SLAM graphs $\{G_{\epochidset_{1}}, \dots,  G_{\epochidset_{n-1}} \}$.
Each graph contains data from one or more epochs (via map merging), represented by $\epochidset$.
We denote the current epoch ($\epochid^\ast$) and sets that contain the current epoch ($\epochid^\ast \in \epochidset^\ast$) with an asterisk.
$G_{\epochidset^\ast_{n}}$ is the graph that is actively used by SLAM, meaning that the robot is localized wrt. to this graph, and data from the current epoch is added to it.
When an epoch ends, the active graph gets moved into the set of inactive graphs and a new empty active graph $G_{\epochidset^\ast_{n+1}}$ is created when a new epoch starts.
Note that a graph implicitly defines a map, so, in the following, we use the two terms interchangeably.

\subsection{Map Merging}
\begin{figure*}
	\centering
	\begin{subfigure}[b]{.31\linewidth}
		\includegraphics[width=\linewidth]{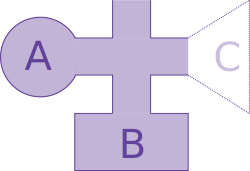}
		\caption{The inactive map.}
		\label{fig:toy_example_inactive_map}
	\end{subfigure}
	\quad
	\begin{subfigure}[b]{.31\linewidth}
		\includegraphics[width=\linewidth]{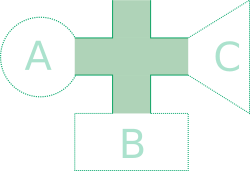}
		\caption{The active map.}
		\label{fig:toy_example_active_map}
	\end{subfigure}
	\quad
	\begin{subfigure}[b]{.31\linewidth}
		\includegraphics[width=\linewidth]{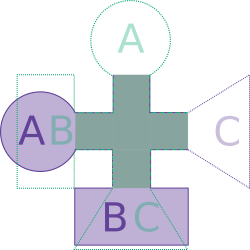}
		\caption{A risky, invalid merge.}
		\label{fig:toy_example_risky_merge}
	\end{subfigure}
	\\
	\vspace{3mm}
	\begin{subfigure}{.31\linewidth}
		\includegraphics[width=\linewidth]{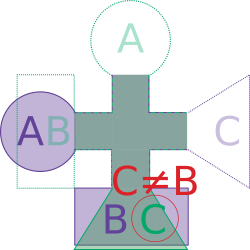}
		\caption{Detectable via \emph{unexpected appearance}.}
		\label{fig:toy_example_unexpected_appearance}
	\end{subfigure}
	\quad
	\begin{subfigure}{.31\linewidth}
		\includegraphics[width=\linewidth]{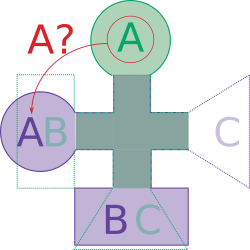}
		\caption{Detectable via \emph{better hypothesis}.}
		\label{fig:toy_example_better_hypothesis}
	\end{subfigure}
	\quad
	\begin{subfigure}{.31\linewidth}
		\includegraphics[width=\linewidth]{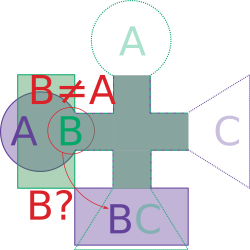}
		\caption{Detectable both ways.}
		\label{fig:toy_example_both_cases}
	\end{subfigure}
	\caption{
		Toy environment for illustrating the two types of approaches for detecting invalid merges.
		It consists of three different rooms A, B, and C, connected by a four-way crossing.
		The inactive map is visualized in blue (\subref{fig:toy_example_inactive_map}), the active map in green (\subref{fig:toy_example_active_map}).
		Unexplored parts of the environment are drawn with dotted lines.
		In (\subref{fig:toy_example_risky_merge}), a risky merge is performed, which turns out to be wrong (rotated clockwise by \SI{90}{\degree}).
		At the time of the merge, however, the merged map appears to be consistent.
		Depending on which path the robot takes, we can detect that the merge was wrong via \emph{unexpected appearance} (red not equal sign) (\subref{fig:toy_example_unexpected_appearance}), \emph{better hypothesis} (red questionmark) (\subref{fig:toy_example_better_hypothesis}), or by both classes of methods (\subref{fig:toy_example_both_cases}).
	}
	\label{fig:toy_example}
\end{figure*}

In order to know which maps to merge, we perform place recognition to find the relative pose between two graphs.
For this purpose, one can for example use RANSAC approaches~\cite{Fontanelli2007}, an IRON key point matcher \cite{Schmiedel2015}, a FLIRT keypoint matcher \cite{Tipaldi2010}, or similar approaches.
Once a unique match has been found, we assume that the algorithm provides a set $E_\text{loop}$ of one or more loop closure edges that connect vertices of the two graphs.

We assume that the active map is given by a SLAM graph $G_{\epochidset^{\ast}_{1}} = (V_{\epochidset^{\ast}_{1}}, E_{\epochidset^{\ast}_{1}})$ with a set of vertices $V_{\epochidset^{\ast}_{1}}$ and a set of edges $E_{\epochidset^{\ast}_{1}}$.
Correspondingly, the inactive map is given by $G_{\epochidset_{2}} = (V_{\epochidset_{2}}, E_{\epochidset_{2}})$.
To construct the merged graph, we first create a new graph based on the union of both graphs and the loop closure edges, i.e., $G_{\epochidset^{\ast}_{1} \cup \epochidset_{2}} = (V_{\epochidset^{\ast}_1} \cup V_{\epochidset_2}, E_{\epochidset^{\ast}_1} \cup E_{\epochidset_2} \cup E_\text{loop})$ where $E_\text{loop} \subseteq V_{\epochidset^{\ast}_{1}} \times V_{\epochidset_{2}}$.
Graph optimization will automatically align the graphs into a single coordinate frame, or, alternatively, the graphs can be manually transformed based on the relative pose(s) in $E_\text{loop}$ to give better initial values for optimization. 
After merging two maps, it makes sense to try to find additional loop closures where the maps overlap to improve their alignment.

During live operation, we only search for matches between all inactive graphs $G_{\epochidset_i} \in \left\{G_1, \dots, G_{\epochidset_{n-1}}\right\}$ and $G_{\epochidset^{\ast}_{n}}$, but not between different inactive maps.
When merging, we make sure that we continue to use $G_{\epochidset^{\ast}_{n}}$'s reference frame to avoid significant pose jumps, which makes map merges much more pleasant for other navigation modules.
We continue with this procedure, as long as there are inactive maps left.
So, there may be several subsequent merges of several different inactive maps into the same active map.
Furthermore, note that an inactive map $G_{\epochidset_i}$ may also have been created as a result of merges themselves ($\left\|\epochidset_i\right\| > 1$).

\subsection{Unmerging}

After merging an inactive map into the active map $G_{\epochidset^{\ast}_n}$, we run our invalid merge detection (see Sec.~\ref{sec:invalidmergedetection}) whenever a new vertex is added, i.e., new sensor information becomes available.
For these algorithms, it is crucial to differentiate vertices of the current epoch $V_{\left\{\epochid^\ast\right\}} \subseteq V_{\epochidset^{\ast}_n}$ from vertices that stem from other epochs $V_{\epochidset^\ast_n \setminus \left\{ \epochid^\ast \right\} } \subset V_{\epochidset^{\ast}_n}$. 
In case we detect an invalid merge, we undo \emph{all} merges into the active map, such that $\epochidset^{\ast}_n = \left\{\epochid^\ast\right\}$.
In principle, this may lead to undoing more merges than necessary in some cases, but reduces the complexity of the problem.
The benefit of more selectively undoing merges seems limited in most use-cases, where multiple subsequent merges are the exception.

Before we merge an inactive map into the active map, we create a backup of that inactive map and store it in RAM or on disk, depending on the available memory.
If lidar scans are never modified, it is sufficient to only backup the graph structure without duplicating the scans.
Once we want to undo a merge, we simply restore the inactive map(s) from backup.

As far as the active map is concerned, the exact solution would require to also create a backup before the merge, store all sensor data after the merge, and reprocessing it after restoring the backup.
In practice, this is very slow and requires a lot of memory.
For this reason, we propose an alternative solution:
We remove all vertices $v \in V_{\epochidset^{\ast}_n \setminus \left\{ \epochid^\ast \right\} }$ from the active map that were merged into it.
In addition, we remove all edges connected to these vertices.
Finally, we optimize the graph to get a good estimate of $G_{\left\{\epochid^\ast\right\}}$.
This is not guaranteed to lead to exactly the same result as if the merge had never taken place.
However, we did not find any cases where this leads to significantly worse results.

If graph pruning such as \cite{Kurz2021} is used, special care needs to be taken.
If an inactive graph is merged into the active graph, we cannot prune information from the active graph because the same area is covered in the merged graph as this merge might later be undone, which would break the active graph.
Therefore, we perform pruning on the active SLAM graph only considering vertices from the current epoch.
As soon as epoch $\epochid^\ast$ ends, and we did not find an invalid merge, we can prune the complete graph.

\section{Invalid Merge Detection}
\label{sec:invalidmergedetection}
\begin{figure*}
	\centering
	\begin{subfigure}{0.31\linewidth}
		\fbox{
		\includegraphics[width=.9\linewidth]{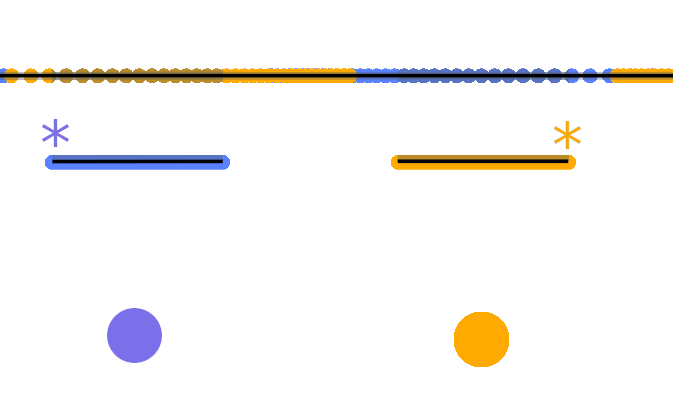}
		}
		\caption{Input data: $D^t$~\textcolor[rgb]{0.482,0.439,0.914}{$\blacksquare$} and $D^s$~\textcolor[rgb]{1.0,0.667,0.0}{$\blacksquare$}.}
    \label{fig:pcp_comp_pts}
	\end{subfigure}
	\quad
	\begin{subfigure}{0.31\linewidth}
		\fbox{
		\includegraphics[width=.9\linewidth]{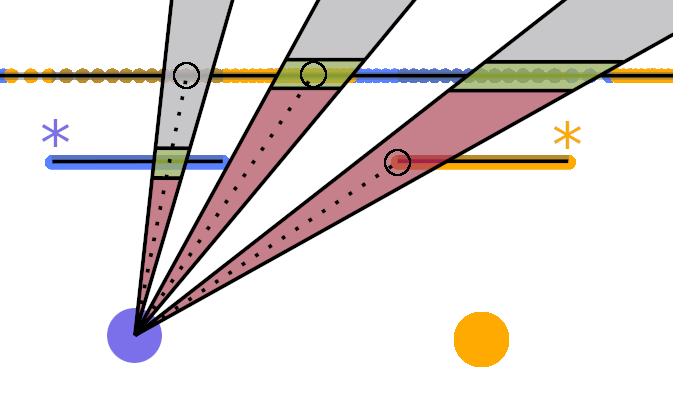}
		}
		\caption{Example rays and classification regions.}
    \label{fig:pcp_comp_rays}
	\end{subfigure}
	\quad
	\begin{subfigure}{0.31\linewidth}
		\fbox{
		\includegraphics[width=.9\linewidth]{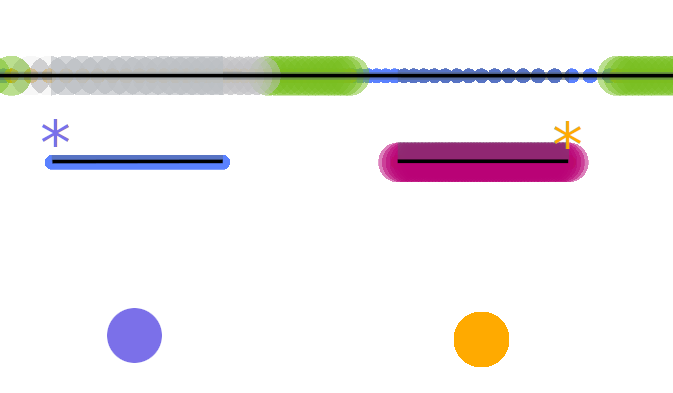}
		}
		\caption{Final classification result.}
    \label{fig:pcp_comp_cls}
	\end{subfigure}
  \caption{
    Visualization of a single scan pair comparison with adapted algorithm from~\cite{Underwood2013}.
    Sensor origins are big dots, sensor measurements are small dots.
    The long wall is perceivable by both scans, but the left/right wall (\textcolor[rgb]{0.482,0.439,0.914}{*}/\textcolor[rgb]{1.0,0.667,0.0}{*}) only by $D^t$/$D^s$, respectively.
    Red \textcolor[rgb]{0.698,0.145,0.298}{$\blacksquare$} corresponds to \emph{change}, green \textcolor[rgb]{0.596,0.686,0.239}{$\blacksquare$} to \emph{agree}, and grey \textcolor[rgb]{0.714,0.714,0.722}{$\blacksquare$} to \emph{no info}.
    Note that only $D^s$ is getting classified.
  }
  \label{fig:change_detection_classes}
\end{figure*}

We differentiate two different types of approaches that can detect invalid map merges, \emph{better hypothesis} and \emph{unexpected appearance}. The first one detects that later a better merge hypothesis is found than the one used during the merge and the second one detects that the observed environment contradicts the environment in the merged map. 
In this context, appearance can refer to any observable property of the environment such as its geometry or its color.
They are illustrated using a toy example environment in Fig.~\ref{fig:toy_example}.
While Fig.~\ref{fig:toy_example_unexpected_appearance} and Fig.~\ref{fig:toy_example_better_hypothesis} show cases where only one type of approaches can detect the invalid merge, this is unusual in practice.
In both cases, the other type of approach would likely become usable as well if the robot continues to explore its environment.

The first approach finds a \emph{better hypothesis} of how two maps can be merged via place recognition.
Assume the same place is recorded in both maps.
If a merge is correct, this place should be located at a single location in the merged map.
According to an incorrectly merges map, a place can appear to exist at two different locations (see Fig.~\ref{fig:toy_example_better_hypothesis} and Fig.~\ref{fig:toy_example_both_cases}).
When detecting invalid merges by generating better hypotheses, there are two failures modes:
First, due to perceptual aliasing, two different places A and A' can look similar.
This can cause these methods to undo a correct merge.
Second, changes in appearance of room A can cause failure to detect the invalid merge in Fig.~\ref{fig:toy_example_better_hypothesis}.

The second approach detects an \emph{unexpected appearance} of the environment, e.g., via methods that can predict map quality or differences in maps.
An invalid merge can cause different places to overlap in the merged map (see Fig.~\ref{fig:toy_example_unexpected_appearance} and Fig.~\ref{fig:toy_example_both_cases}).
This usually leads to walls intersecting at unusual angles and is often easy to spot for humans as a ``broken map''.
Note that in order to detect \emph{unexpected appearance}, it is not necessary to recognize a place: ``This is place C, therefore it cannot be place B''.
Instead, finding that ``this is not place B'' is sufficient.
This fact suggests that detecting \emph{unexpected appearance} might be the easier problem to solve.
The two failure modes for this approach are the other way around:
Changes in appearance of a place can cause these methods to undo correct merges and perceptual aliasing can cause them to miss a invalid merge.

In the following, we focus only on methods for detecting \emph{unexpected appearance}.
An advantage of this type of methods is that it can run on local information around the robot's current location.
In contrast, when searching for a better hypothesis, every new bit of sensor data needs to be compared to all known places.
For this, every algorithm can be used that is capable of finding a merge to begin with.
In principle, using both types of methods in parallel will likely yield better results at increased runtime costs.

\subsection{Change Detection}
The following approach is based on a method from~\cite{Underwood2013}, for detecting changes in 3D point clouds.
It operates on a pairs of scans $(D^t, D^s)$.
Each scan consists of $D^a = (x^a, S^a)$, where $x^a$ is the scan's global 6D pose and $S^a$ is the set of endpoints in spherical (3D scans) or polar (2D scans) coordinates.
The algorithm returns a subset of point indices $C^t = \left\{ i_1, \cdots, i_M \right\}, i \in \left\{ 1, \cdots, \left| D^t \right| \right\} $ that violate the free-space of $D^s$.
When $(D^t, D^s)$ are aligned correctly, the points in $C^t$ most likely depict dynamic or semi-static objects.
But aligning $(D^t, D^s)$ incorrectly can also cause a lot of free-space violations, e.g. by criss-crossing wall segments.
This fact can be used to detect invalid merges.

We adapt the algorithm from~\cite{Underwood2013} such that it additionally differentiates whether a measurement in $D^t$ agrees with how $D^s$ has perceived the environment (green \textcolor[rgb]{0.596,0.686,0.239}{$\blacksquare$} points in Fig.~\ref{fig:pcp_comp_cls}) or $D^s$ has no information about that part of the environment (grey \textcolor[rgb]{0.714,0.714,0.722}{$\blacksquare$} points in \ref{fig:pcp_comp_cls}).
For that, we adapt the range check (see \cite[Algorithm~1, line 4]{Underwood2013}) such that it returns a list of class labels $\left( c^1, \cdots, c^{\left| D^t \right|} \right)$, with
\begin{align*}
  c^p = \begin{cases}
    \text{change}, &r^t < r^s_\text{min} + \rangetolerance \\
    \text{agree}, &\begin{aligned}& r^t \in [r^s_\text{min} - \rangetolerance, r^s_\text{max} + \rangetolerance]\\
                                 & \land r^s_\text{min} + r^s_\text{max} \leq 2 \cdot \rangetolerance \end{aligned}\\
    \text{no info}, &\text{otherwise}
  \end{cases} 
\end{align*}
depending on the min and max range $r^s_\text{min}$ and $r^s_\text{max}$ of nearby points from $S^s_\text{near} \subset S^s$.
The \emph{agree} interval can become large when $S^s_\text{near}$ is looking at things at a  steep angle or at object boundaries.
To address this issue, we additionally check that this interval is smaller than $2\cdot\rangetolerance$.
In our experiments, we use $\rangetolerance=\SI{0.1}{\meter}$ and $\neighborhoodslicesize=\SI{3}{\degree}$ (for calculating $S^s_\text{near} \subset S^s$).

Whenever a new scan $D^i$ is added to a map $G_\epochidset^\ast, \left\| \epochidset^\ast \right\| > 1$ after a merge, we take the most recent $\nkeyframes \in \mathbb{N}$ scans from the current epoch (see Fig.~\ref{fig:intro_example}, vertices highlighted in blue~\textcolor[rgb]{0.00,0.56,0.81}{$\blacksquare$}).
Then, we pair each of these scans with all scans from other epochs, that overlap by at least $\minoverlapthreshold$ meters (see Fig.~\ref{fig:intro_example}, vertices highlighted in gray~\textcolor[rgb]{0.32,0.37,0.41}{$\blacksquare$}).
For this, we utilize a visibility grid, where each cell contains the ids of all scans from other epochs that have observed it.
Since the grid does not contain any data from the active epoch $\epochid^\ast$, it is sufficient to compute it once after the merge.

Our adapted algorithm gets applied to all paired scans symmetrically.
Each point will get multiple classifications, one for each paired scan.
Additionally, new classifications $c^p_\text{new}$ will occur over time, whenever new scans are submitted into the map.
For each point, we update and store the class label $c^p_\text{fused}$ using the following fusion scheme:
\begin{align*}
  c^p_\text{fused} = \begin{cases}
    \text{agree}, &c^p_\text{fused} = \text{agree} \lor c^p_\text{new} = \text{agree} \\
    \text{no info}, &c^p_\text{fused} = \text{no info} \land c^p_\text{new} = \text{no info} \\
    \text{change}, &\text{otherwise}
  \end{cases} 
\end{align*}
This scheme prioritizes the \emph{agree} class over the \emph{change} class in order to become robust, e.g., against obstacles that appear see-through to the lidar.
We cache the fused classification results during a SLAM session.
However, when the estimated pose of a vertex has changed significantly during graph optimization, e.g., due to loop closures, the cache gets invalidated and the classifications are re-done if the vertex is still relevant for invalid merge detection.

Finally, we count how often the different number of classes occur for all considered points and calculate the ratio $\invalidityratio = \frac{C^\text{change}}{C^\text{agree} + C^\text{change}} \in [0,1]$ and perform an unmerge if it is above a threshold $\invalidityratio > \invalidityratiothreshold$.
Note that points with no information do not influence $\invalidityratio$.

\subsection{Gridmap}
\label{sec:gridmap}
\begin{figure*}
	\centering
  \includegraphics[width=\linewidth]{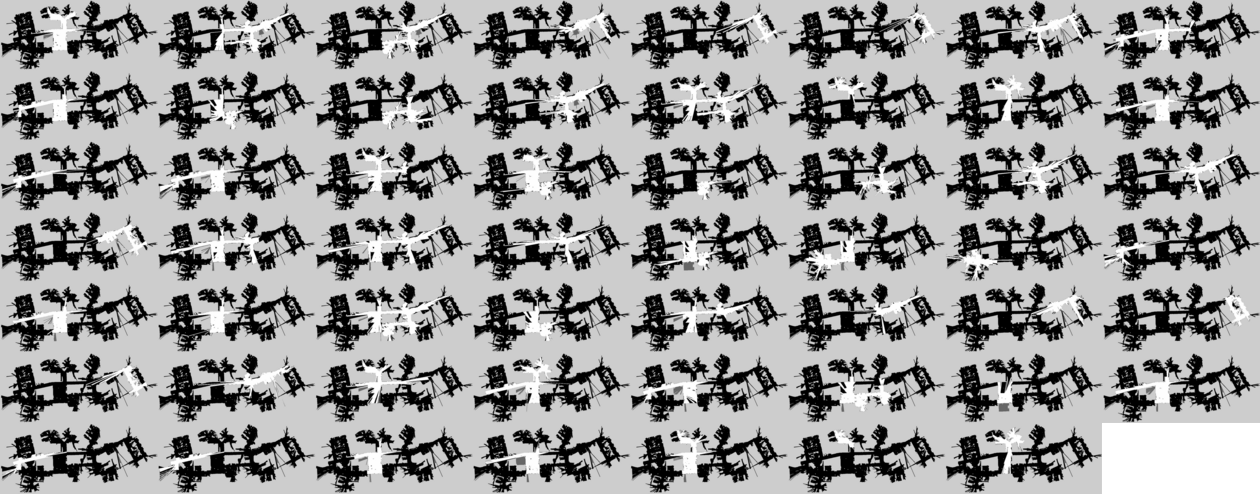}
  \caption{Maps from the segments of the MIT Stata Center dataset (white), aligned to a reference map created from ``2012-02-02-10-44-08'' (black).}
  \label{fig:segments_mit_stata}
\end{figure*}

In the following, we propose a gridmap-based approach to detect invalid merges.
For this purpose we consider a window of the $\nkeyframes \in \mathbb{N}$ most recent keyframes.
First, we compute the axis-aligned bounding box of all scans associated to these keyframes.
This bounding box corresponds to the area of interest.

Then, we compute two gridmaps in this area; the first gridmap is based on the newest keyframes and the second gridmap is based on vertices whose epoch $\epochid$ is different from the current epoch.
i.e., that have been merged into the active map.
Our gridmap has three possible states for each cell: empty, occupied, and unknown. In our experiments, we use a cell size of \SI{2.5}{\centi\meter}.

After computing the gridmaps, we perform a cell-wise comparison. 
Specifically, for cells $a$, $b$ from the first and second gridmap respectively, we determine the sets
\begin{align*}
C_\text{overlap} =& \{ \neg  a \text{ is unknown} \land \neg b \text{ is unknown} \} \\
C_\text{contradictions} =& \{ (a \text{ is empty} \land b \text{ is occupied}) \\
 &\lor (a \text{ is occupied} \land b \text{ is empty} \}
\end{align*}
and compute the invalidity ratio
\begin{align*}
r = \begin{cases}
|C_\text{overlap}| \geq \minoverlapthreshold, & \frac{|C_\text{contradictions}| }{|C_\text{overlap}|} \\
|C_\text{overlap}| < \minoverlapthreshold, &  0 
 \end{cases} 
\end{align*}
for a minimum overlap threshold $\minoverlapthreshold \geq 0$.
It holds that $r \in [0,1]$ and larger $r$ indicates more contradictions and thus, a higher probability that an invalid merge occurred.

Because occupied areas such as walls tend to be quite thin in the gridmap, a slight misalignment by a single cell already leads to a contradiction.
Therefore, we use the dilation operation (with a square structuring element of $7\times7$ cells) on both gridmaps before comparing them to increase the size of occupied areas.

\subsection{Map Quality Measures -- Entropy}
Another idea to detect invalid merges is to use some kind of map quality metric in order to assess the quality of the map and to use a low quality as an indicator that a invalid merge has happened.
Different map quality metrics have been proposed in literature, e.g., \cite{Almqvist2017}, \cite{Filatov2017}.
Some of these metrics measure aspects of the map's quality that are not suitable for detecting invalid merges since they are designed to respond to noisy maps or slightly misaligned scans.
As an example of a map quality approach, we implemented the entropy measure proposed by~\cite{Droeschel2018}, \cite[Sec.~VI]{Droeschel2014}.

We compute the entropy based on the point cloud obtained by combining all lidar scans of the SLAM map. The entropy of a point $q_k \in \{q_1, \dots, q_Q\}$ is given by 
\begin{align*}
	h(q_k) = \frac{1}{2} \ln \left( \det( 2\pi e \Sigma(q_k) ) \right)
\end{align*}
where $\Sigma(q_k)$ is the covariance of all points in a radius $r= \SI{0.3}{\meter}$ around $q_k$.
This corresponds to the differential entropy of a Gaussian distribution with covariance $\Sigma(q_k)$.
The entropy of the entire map is obtained by averaging over all points
\begin{align*}
	H = \frac{1}{Q} \sum\nolimits_{k=1}^Q h(q_k) \ .
\end{align*}
Because we are not interested in the absolute entropy of the map but rather of the influence of the merges, we compute 
\begin{align*}
	\Delta H = H_\text{all} - \frac{1}{2} (H_\text{current epoch} - H_\text{other epochs}) \in \mathbb{R} \ .
\end{align*}
This way, we can detect if the merge increases or reduces the entropy and interpret an increase in entropy as an indicator of an invalid merge.

\subsection{Loop Closure Verification -- Histogram}
Another related problem to invalid merge detection is loop closure verification.
Thus, we implemented the histogram-based approach proposed by from Corso~\cite[Sec.
  3.1]{Corso2013}, which is in turn inspired by~\cite{Barla2003}.
This method somewhat resembles the gridmap approach from Sec.~\ref{sec:gridmap}.
The key difference is that does not consider each grid cell as occupied or unoccupied but instead counts the number of points in each grid cell to obtain a histogram.
Corso uses a normalized histogram in his paper, but we decided to use an unnormalized histogram to be able to compare maps with vastly different sizes and thus different numbers of points.
Specifically, we compute the unnormalized histograms for the most recent $n$ vertices of the current epoch $h_1(\cdot, \cdot)$ and the unnormalized histogram of all other epochs $h_2(\cdot,\cdot)$.
These two histograms can be compared using the intersection kernel with subsequent normalization
\begin{align*}
	c = \frac{\sum_{i,j} \min\left( h_1(i,j), h_2(i,j) \right)}{\sum_{i,j} h_1(i,j)} \in [0,1]
\end{align*}
where $h_1(i,j)$ and $h_2(i,j)$ are the number of points in the grid cell at location $(i,j)$ in the respective map.
We then use $1-c \in [0,1]$ as an invalidity measure since large $c$ indicates a good agreement between $h_1(\cdot, \cdot)$ and $h_2(\cdot, \cdot)$ whereas small $c$ suggests a mismatch between the two histograms. In our experiments, we use a histogram cell size of \SI{0.5}{\meter}.

\section{Evaluation}
\begin{table}
  \centering
  \caption{Datasets used for testdata sequence generation.}
  \label{tab:datasets}
  \begin{tabular}{ llrr }
    \toprule
    \textbf{Dataset}      & \textbf{Recording}  & \textbf{Segments} & \textbf{Duration} \\
                          &                     &                   & (minutes) \\
    \midrule
    \multirow{9}{*}{\shortstack[l]{\textbf{Flats and}\\ \textbf{Offices}}}
                          & Office 1            & 33                & 242 \\ 
                          & Office 2            & 20                & 115 \\ 
                          & Flat 1.1            & 14                & 103 \\ 
                          & Flat 1.2            & 12                & 47  \\ 
                          & Flat 2              & 35                & 129 \\ 
                          & Flat 3              & 13                & 109 \\ 
                          & Flat 4              & 7                 & 28  \\ 
                          & Flat 5              & 9                 & 39  \\ 
    \cmidrule{2-4}           & Sum                 & 143               & 812 \\
    \midrule
    \multirow{5}{*}{\shortstack[l]{\textbf{MIT Stata}\\ \textbf{Center}}}
                          & 2012-01-18-09-09-07 & 18                 & 36  \\
                          & 2012-01-25-12-14-25 & 11                 & 20  \\
                          & 2012-01-25-12-33-29 & 7                  & 14  \\
                          & 2012-01-28-11-12-01 & 19                 & 36  \\
    \cmidrule{2-4}           & Sum                 & 55                 & 106 \\
    \bottomrule
  \end{tabular}
\end{table}
\begin{table}
  \centering
  \caption{Testdata sequences used for evaluation.}
  \label{tab:sequences}
  \begin{tabular}{ lrrr }
    \toprule
     & & \textbf{Correct} & \textbf{Invalid} \\
    \textbf{Dataset}           & \textbf{Sequences} & \textbf{Merges} & \textbf{Merges} \\
    \midrule
    \textbf{Flats and Offices} & 436                & 252                     & 184 \\
    \textbf{MIT Stata Center}  & 1114               & 559                     & 555 \\
    \bottomrule
  \end{tabular}
\end{table}

Our evaluation tests how well the approaches described in Section~\ref{sec:invalidmergedetection} can differentiate correct merges from invalid merges.
In addition, we investigate the computation time they require.
We use two datasets, an internal ``Flats and Offices'' dataset and the public ``MIT Stata Center'' dataset\footnote{\url{https://projects.csail.mit.edu/stata/}}.

\subsection{Evaluation Scheme}
We run our SLAM pipeline on a set of sequences with two segments each.
Each segment contains a small part of a dataset and corresponds to a single epoch $\epochid$ (see Sec.~\ref{sec:integration}).
The contents of the first segment are simply used to generate a map $G_{\epochidset_1}$.
When the second segment starts, our SLAM pipeline creates a new map $G_{\epochidset_2}$, unconnected to the first one.
Usually, we would automatically recognize places in $G_{\epochidset_1}$ and perform a map merge (if the maps overlap).
For this evaluation, however, we deactivate automatic map merges and force a map merge with a predefined pose at a predefined point in time, which may be correct or incorrect.
This approach ensures that we get consistent merges for evaluation (place recognition might be randomized or gets improved) and enables us to manually add specific merges.

After the merge, all invalid merge detectors output the respective invalidity ratio $\invalidityratio$ for each new scan but, during evaluation, we do \emph{not} undo invalid merges.
This allows us to retrieve the maximum invalidity ratio that each approach reaches for each test sequence.
If this maximum invalidity ratio lies above a threshold $\invalidityratio >= \invalidityratiothreshold$, the maps would have gotten unmerged at some point during processing.
With this scheme, we can calculate the number of true positives, false positives, true negatives, and false negatives for all possible $\invalidityratiothreshold$.
While this scheme does not explicitly consider how quickly an invalid merge gets detected and undone, only that it happens at some point before the second segment is finished, it is useful for comparing classification performance and finding a good $\invalidityratiothreshold$.

\subsection{Dataset: MIT Stata Center}
The MIT Stata center dataset features indoor scenes from an academic building, from which we only use the subset depicted in Table~\ref{tab:datasets}.
The chosen recordings are all from the second floor of the building and start and stop roughly at the same location.
We use the Hokuyo UTM-30LX Laser data from the base of the PR2 robot, the robot's raw wheel odometry, and the Microstrain 3DM-GX2 IMU.
In order get more evaluation sequences from the same amount of data, we segment each run into smaller pieces of about \SI{100}{\second}.
With this approach, we get a lot of segments that are still long enough for place recognition and invalid merge detection, and with different start and stop poses.

The sequences for our evaluation with this dataset were generated with minimal manual labeling required:
First, we create a reference map from the unsegmented ``2012-02-02-10-44-08'' dataset.
Then, we create maps for all $55$ segments and align them to the reference map via our place recognition module.
To make sure these transformations are correct, we inspect and correct them manually, if necessary.
An overview of the segments aligned to the reference map is shown in Fig.~\ref{fig:segments_mit_stata}.
Using this information, we can calculate the relative transformations between all $2970 = 55\cdot54$ pairs of segments.
For each such pair, we estimate the overlap of their maps and discard it, if it is below \SI{1.5}{\square\meter}.
The remaining pairs form the set of evaluation sequences.
The last step is getting a map merge pose and whether that pose corresponds to a correct or an invalid merge.
For this, we run the sequence in our pipeline with extremely careless parameters for the place recognition module.
This ensures that a merge takes place in almost all cases, and that we get invalid merges frequently.
Finally, we can differentiate correct merges from invalid merges using the relative transformation estimated via the reference map.

\subsection{Dataset: Flats and Offices}
Our internal Flats and Offices dataset consists of recordings in two office spaces and five different flats, see the upper part of Table~\ref{tab:datasets}.
For Flat1, we use data from two floors (Flat1.1 and Flat1.2).
The recordings contain data from a low resolution \SI{360}{\degree} 2D lidar, the robot's wheel odometry and its IMU.
The test sequences (see Table~\ref{tab:sequences}) of Flats and Offices contains a mixture of correct and invalid merges that our SLAM pipeline performed during other tests, as well as manually added merges.
Segments in this dataset often correspond to a run through parts of the flat or office, e.g. one or two rooms.
Some segments also explore the whole environment.
The same part of the environment gets explored in multiple segments, at different points in time.
We can assemble these segments into sequences in many different ways to create Lifelong SLAM scenarios that include incremental map building, changes in the environment, and kidnapping.

\subsection{Results}
\begin{figure}
	\centering
	\fbox{
		\includegraphics[width=.43\linewidth]{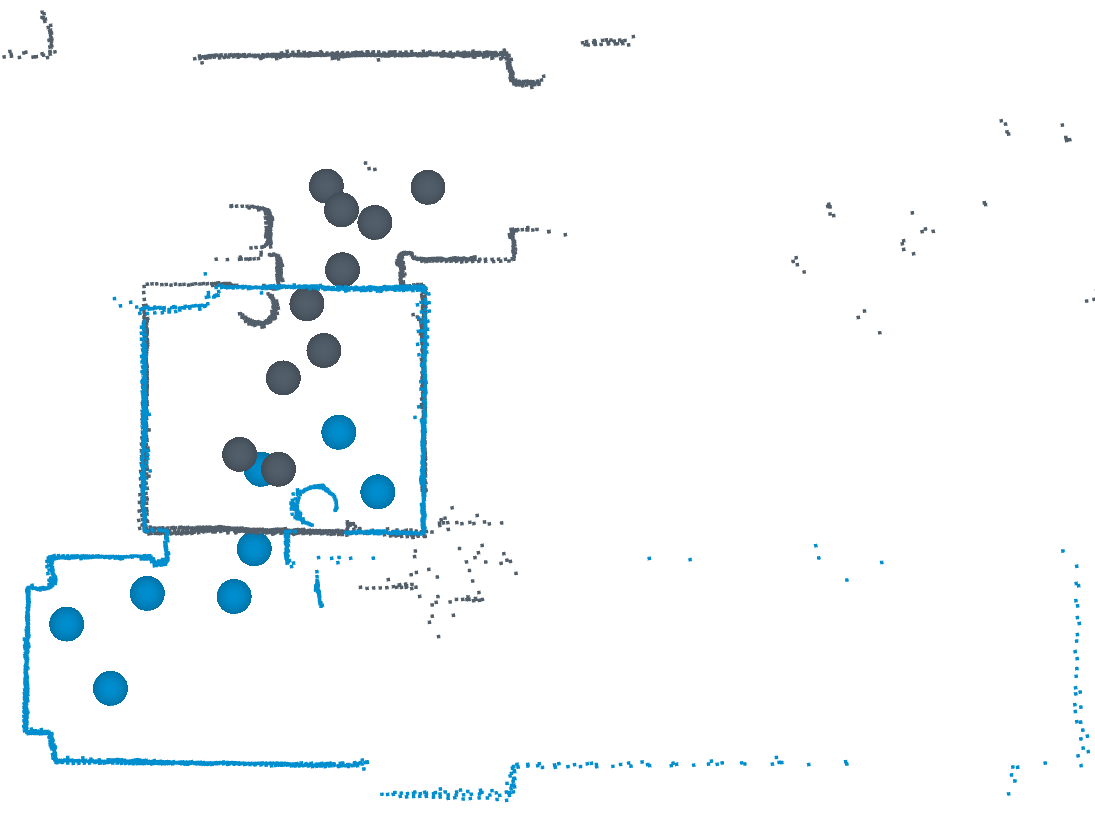}
	}
	\fbox{
		\includegraphics[width=.43\linewidth]{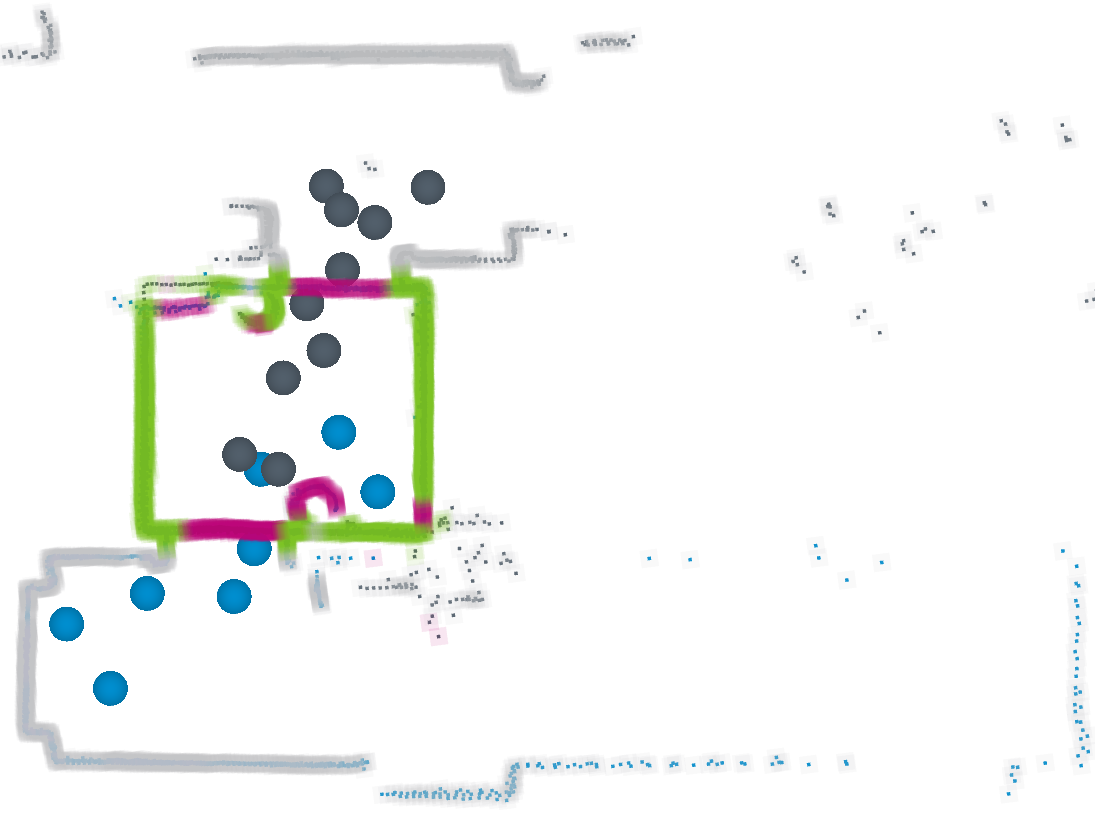}
	}
	\caption{Example of an invalid merge (rotated by \SI{180}{\degree}) that is hard to detect via \emph{unexpected appearance}.}
	\label{fig:failure_cases}
\end{figure}

\begin{figure}
    \centering
    \begin{subfigure}{.49\linewidth}
        \centering
        \includegraphics[width=\linewidth]{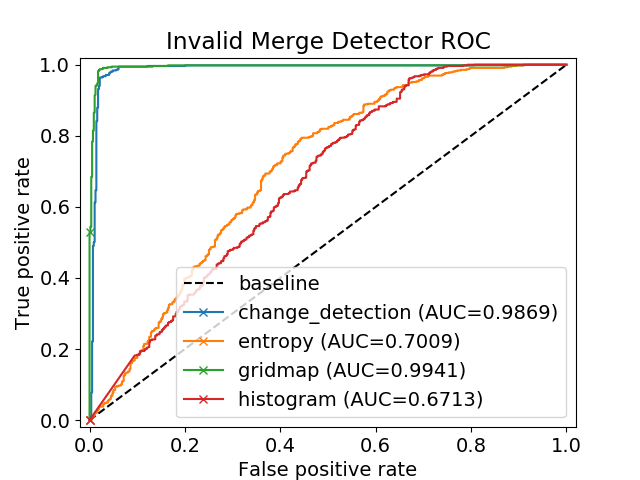}\\
        \includegraphics[width=.9\linewidth]{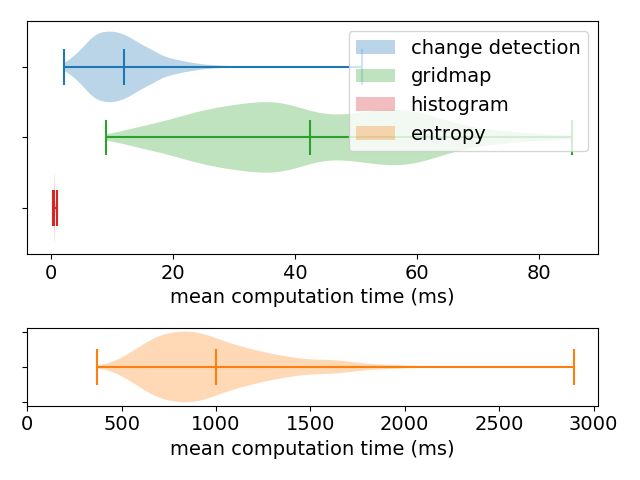}
        \caption{Results for MIT Stata Center.}
        \label{fig:results_mit_stata}
    \end{subfigure}
    \begin{subfigure}{.49\linewidth}
        \centering
        \includegraphics[width=\linewidth]{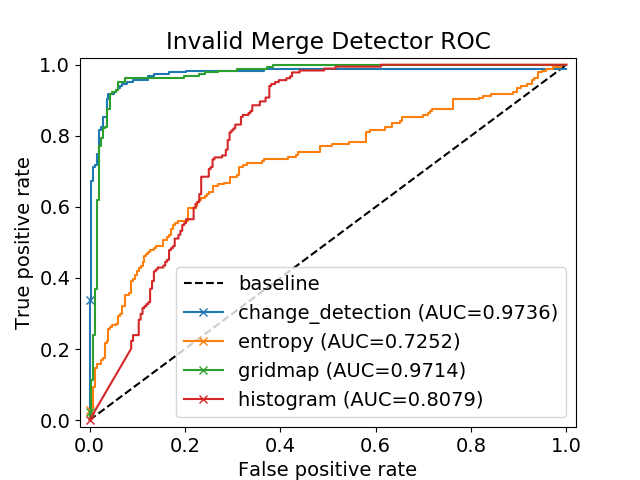}\\
        \includegraphics[width=.9\linewidth]{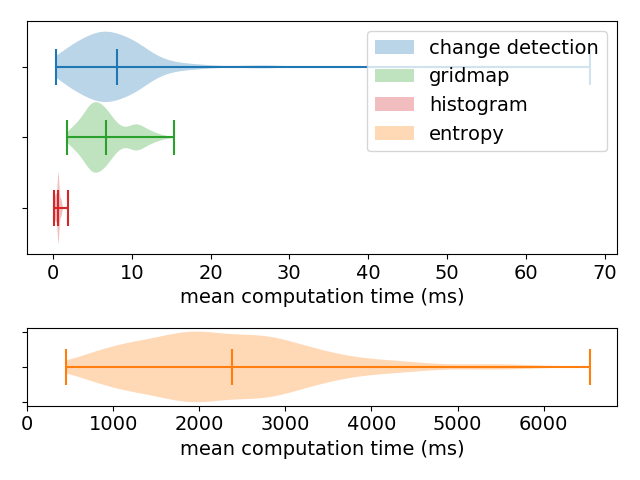}
        \caption{Results for Flats and Offices.}
        \label{fig:results_flats_custom}
  \end{subfigure}
  \caption{
    The top row compares classification performance via ROC curves.
    The bottom row compares runtime costs via violin plots.
    Note that we chose a different scale for the entropy method.
  }
  \label{fig:results}
\end{figure}

\begin{table}
  \centering
  \caption{Evaluation results}
  \label{tab:results}
  \begin{tabular}{ llrr }
    \toprule
    \textbf{Dataset}  & \textbf{Method}           & \textbf{AUC} & \textbf{Mean compute time} \\
                      &                           &              & (ms per new vertex) \\
    \midrule
    \multirow{4}{*}{\shortstack[l]{\textbf{MIT Stata}\\ \textbf{Center}}}
                      & Change Detection          & 0.987        & 11.87    \\
                      & Gridmap                   & \textbf{0.994}        & 42.38    \\
                      & Entropy                   & 0.701        & 998.40   \\
                      & Histogram                 & 0.671        & 0.518    \\
    \midrule
    \multirow{4}{*}{\shortstack[l]{\textbf{Flats and}\\ \textbf{Offices}}}
                      & Change Detection          & \textbf{0.974}        & 8.10     \\
                      & Gridmap                   & 0.971        & 6.75     \\
                      & Entropy                   & 0.725        & 2376.48  \\
                      & Histogram                 & 0.808        & 0.700    \\
    \bottomrule
  \end{tabular}
\end{table}

Our evaluation results are shown in Fig.~\ref{fig:results} and Table~\ref{tab:results}.
We use Receiver Operating Characteristic (ROC) curves in order to compare the classification performance of our proposed methods.
Regarding the compute time, note that we run each method whenever a new vertex gets added to our graph, but only if a a merge has already occurred.
We store the runtime for each of these function calls and show the mean computation time over all evaluated sequences.
All presented results were obtained on a laptop with a single core of an Intel\textregistered~i7-8650U~@~1.90GHz CPU.

The evaluation shows that the change detection and the gridmap methods perform similarly.
Depending on the structure of the input data, their runtimes and classification performance differ slightly.
The entropy method and the histogram method have quite poor classification performance in comparison.
The histogram method is the fastest method we considered because it uses a fairly coarse grid.
In contrast, the entropy method is by far the slowest because it needs to compute the neighborhood of every single point in the map, which is extremely costly even though we implemented this step using a k-d tree to improve performance.

Differences in timing between the two datasets are due to differences in the lidar sensors and the length of the segments.
The lidar used in Flats and Offices only provides 360 points per scan, while the Hokuyo lidar in MIT~Stata~Center provides 1040 points per scan and has a longer maximum range.
On the other hand, the MIT Stata Center dataset has smaller segments than Flats and Offices, resulting in maps that have fewer vertices.
In Flats and Offices, there are areas where the number of overlapping scans in the map is quite high.
This explains why change detection (runtime quadratic wrt. considered scans) runs slower than gridmap (runtime linear wrt. considered scans) on Flats and Offices, but is faster on MIT Stata Center.

Especially in the Flats and Offices dataset, there are a few test cases that are arguably undetectable via \emph{unexpected appearance}.
One such example is visualized in Fig.~\ref{fig:failure_cases}.
In order to detect invalid merges like these, we would need an algorithm that continuously searches for \emph{better hypotheses}.

\section{Conclusion}
We have proposed four methods for detecting invalid map merges via \emph{unexpected appearance}.
When early online map merges are needed in a SLAM system, our evaluation shows that the change detection and gridmap methods are capable of significantly improving robustness, without requiring too many resources. 
Note that in case of a valid merge, the compute invested into the change-detection-based approach is not in vain.
It provides information about how semi-static objects changed since the environment was last explored.

However, the current methods for invalid merge detection are not yet flawless.
One promising way to reduce the number of false classifications is to employ methods for detecting \emph{better hypotheses} in parallel, e.g., continuously performing place recognition w.r.t. already merged maps.

The proposed methods for detecting \emph{unexpected appearance} in maps are applicable to other situations besides detecting invalid merges, when the alignment of two sets of scans needs to be verified.
For example, when detecting wrong global loop closures, the two sets consist of scans around the two vertices connected by the global loop closure edge.
With some adaptations, they could also be used a measure for map quality.

Additionally, we proposed a way for handling multiple maps in SLAM.
Its usefulness extends far beyond the basic case where robot is turned off, moved, and resumes operation.
For example, the start of a new epoch can also be used as a recovery mechanism to prevent corruption of the active map due to incremental localization failures, e.g., caused by a featureless corridor when using only lidar odometry.

\bibliographystyle{IEEEtran}
\bibliography{../bib/gkbosch}

\end{document}